\documentclass[conference]{IEEEtran}
\IEEEoverridecommandlockouts
\usepackage{cite}
\usepackage{amsmath,amssymb,amsfonts}
\usepackage{algorithmic}
\usepackage{graphicx}
\usepackage{textcomp}
\usepackage{xcolor}
\def\BibTeX{{\rm B\kern-.05em{\sc i\kern-.025em b}\kern-.08em
    T\kern-.1667em\lower.7ex\hbox{E}\kern-.125emX}}
\begin{document}

\title{Student Activity  Recognition in Classroom Environments using Transfer Learning \\
}

\author{\IEEEauthorblockN{Anagha Deshpande}
\IEEEauthorblockA{\textit{School of Electronics and Communication Engineering} \\
\textit{Dr. Vishwanath Karad MIT World Peace University}\\
Pune, India \\
anagha.deshpande@mitwpu.edu.in}
\and
\IEEEauthorblockN{Vedant Deshpande}
\IEEEauthorblockA{\textit{Department of Information Technology}\\
\textit{Pune Institute of Computer Technology}\\
Pune, India \\
vedantd41@gmail.com}
}
\maketitle

\begin{abstract}
The recent advances in artificial intelligence and deep learning facilitate automation in various applications including home automation, smart surveillance systems, and healthcare among others. Human Activity Recognition is one of its emerging applications, which can be implemented in a classroom environment to enhance safety, efficiency, and overall educational quality. This paper proposes a system for detecting and recognizing the activities of students in a classroom environment. The dataset has been structured and recorded by the authors since a standard dataset for this task was not available at the time of this study. Transfer learning, a widely adopted method within the field of deep learning, has proven to be helpful in complex tasks like image and video processing. Pretrained models including VGG-16, ResNet-50, InceptionV3, and Xception are used for feature extraction and classification tasks. Xception achieved an accuracy of 93\%, on the novel classroom dataset, outperforming the other three models in consideration. The system proposed in this study aims to introduce a safer and more productive learning environment for students and educators.

\end{abstract}

\begin{IEEEkeywords}
human activity recognition, transfer learning, classroom, classification, xception
\end{IEEEkeywords}

\section{Introduction}
Human Activity Recognition is a prominently emerging and dynamic field within artificial intelligence, which revolves around comprehending human gestures or movements and discerning the specific activities they represent. HAR can be useful in various areas like Human-Computer Interaction (HCI), entertainment, smart surveillance, Elderly Living, and autonomous driving systems \cite{b1,b2}. The two primary techniques to recognize human activities are handcrafted feature-based depiction and learning-based representation. Handcrafted representations encounter issues of subjectivity and bias due to their reliance on human intuition. They might lack the ability to generalize to unfamiliar data and face difficulties when dealing with complex tasks. Creating these representations can be a time-intensive process, and maintaining them could pose challenges. Deep learning representations excel at capturing intricate patterns in data while eliminating bias through automated feature extraction. They are good at generalizing to new data, adapting to shifting distributions, and facilitating efficient transfer learning. Although inherent feature hierarchies facilitate knowledge of complicated relationships, their interpretability may offer difficulties \cite{b3,b4}. Deep learning approaches like Convolutional Neural Networks (CNNs) \cite{b5}, Long Short Term Memory (LSTMs)\cite{b6} and Deep Belief Networks (DBNs)\cite{b7} can be employed to perform HAR on large and complex datasets. However, the enormous amount of time, resources, and data needed for deep learning model training is a fundamental challenge \cite{b8}. 

Humans are capable of learning innumerable categories throughout their lives with only a small number of samples. It has been suggested that humans develop this skill through gathering information over a certain period and implementing it to learn new things\cite{b9}. Deep learning models can be trained on one classification task and used on another task, with some finetuning if necessary, by adhering to the Transfer Learning concept \cite{b10}. In transfer learning, there are two primary methods: one involves maintaining the pre-trained network while updating weights, and the other entails employing a pre-trained model for feature extraction and application of a classifier\cite{b11}.

Human Activity Recognition can be used for recognizing student activities in the classroom to improve learning outcomes. Classroom activity detection provides insights into teaching methods, allowing for tailored learning, responding to levels of participation, assisting with data-driven decisions, and improving classroom management for a better learning environment. Classroom activity detection has been performed using audio sources as well as multimodal sources viz. vocal and language modalities \cite{b12,b13}. However, there is scope for further research into the use of videos to detect usual and unusual activities performed by students in a classroom. In this study, several pre-trained deep learning models have been explored for identifying classroom activities and utilized on the authors' newly created and recorded classroom dataset.

\section{Related Work}
This section will examine previous research in the domain of activity recognition and transfer learning. When used for human activity identification, handcrafted feature extraction methods such as SURF, HOG, and PCA integrated feature approaches paired with classifiers produce outstanding results. However, these time-consuming tasks need the use of skilled feature detectors as well as complex feature extraction and representation methodologies. Furthermore, they rely too heavily on the data in consideration and are not sufficiently robust\cite{b14,b15,b16}.

Deep learning-based methods to recognize human activity have become popular because of their capacity to extract features from data automatically and identify intricate patterns. 2D Convolutional Neural Networks were used to recognize human behaviors such as fighting and non-fighting, and the results were noteworthy \cite{b17}.CNNs were used to identify running, walking, activities on the assembly line, and activities in the kitchen by utilizing data from mobile sensors. While 2D CNNs yield impressive results, they are not able to simultaneously collect spatial and temporal features\cite{b18}. Spatial and temporal feature extraction is used by 3D CNNs to identify dynamic patterns and provide better context. Consequently, 3D CNNs produced remarkable results on recordings of surveillance from airports\cite{b19}. Simonayan et al. presented ConvNet, a two-stream convolution layer that demonstrated good performance in identifying human actions on a modest amount of training data\cite{b20}. Convolutional layers, in connection with Long Short Term Memory (LSTM) models, are suitable for processing temporal sequences and enhanced performance while avoiding complex feature extraction\cite{b21}. Sheikh et al. trained various deep models on PSRA6, a novel dataset with six action types, and achieved remarkable results in detecting suspicious human activities\cite{b22}.

The problem with deep learning models is that as the task's complexity increases, so do the resources, time, and data required for training. This challenge can be handled by employing Transfer Learning, which entails applying knowledge obtained in one task to another. Deep et al. applied transfer learning alongside CNNs for video-based activity recognition, achieving a remarkable accuracy on the Weizmann dataset\cite{b23}. Sargano et al. constructed an SVM-KNN classifier on the new data and extracted features using a pre-trained architecture. Outperforming state-of-the-art approaches, their proposed strategy produced impressive results on the Weizmann and KTH datasets\cite{b24}.
This study centers on exploring the utilization of transfer learning for activity recognition within the newly created classroom dataset. 

\section{Methodology}


\subsection{Transfer Learning}
Deep Learning models require extensive resources, time, and data to be employed successfully on complex tasks. Figure 1 depicts the concept of transfer learning, which can be used to mitigate the problems associated with deep learning models. Four models were chosen from a variety of pre-trained models that were investigated for use on the classroom activity detection task: VGG-16 \cite{b25}, ResNet-50\cite{b26}, InceptionV3\cite{b27} and Xception\cite{b28}.

\begin{figure}[ht]
    \centering
    \includegraphics[width=0.42\textwidth]{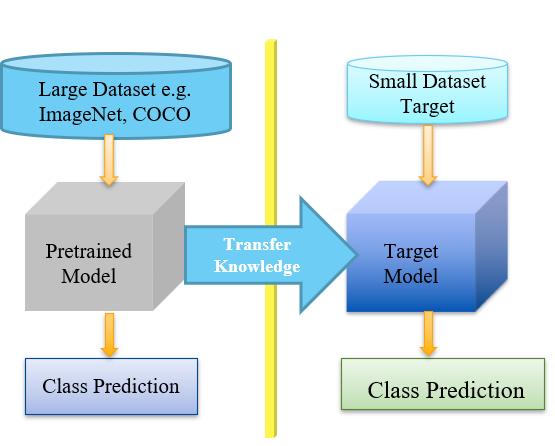}
    \caption{Transfer Learning}
    \label{fig:tf}
\end{figure}

\subsection{Models Used}
VGG-16 is a pre-trained deep learning model that has applications in image classification and face recognition. To proliferate the number of layers and avoid having a large number of parameters, a small 3x3 convolution kernel is used in all layers. VGG takes an RBG image of size 244x244 as an input, which is the average RGB value of images in the training set. Fig. 2 shows the architecture of the VGG-16 model containing 16 layers with 5 sets of convolutional layers, followed by a MaxPool layer.

\begin{figure}[ht]
    \centering
    \includegraphics[width=0.5\textwidth]{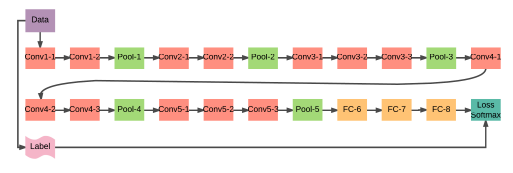}
    \caption{Architecture of VGG-16 model \cite{b29}}
    \label{fig:vgg}
\end{figure}

The winner of the COCO and ILSVRC 2015 competitions, Residual Network (ResNet), makes an effort to solve the difficulty of building very deep neural networks by utilizing residual connections, which helps to handle the vanishing gradient issues and permits the training of deeper models. Fig.3 depicts the architecture of the ResNet-50 model containing fifty layers, with convolutional, pooling, and fully connected layers. When compared to raw feature mappings, skip connections allow the network to understand residual mappings, which improves deep structure optimization.  By utilizing a bottleneck residual block with 1x1 convolutions, ResNet can employ fewer parameters and fewer matrix multiplications.

\begin{figure}[ht]
    \centering
    \includegraphics[width=0.5\textwidth]{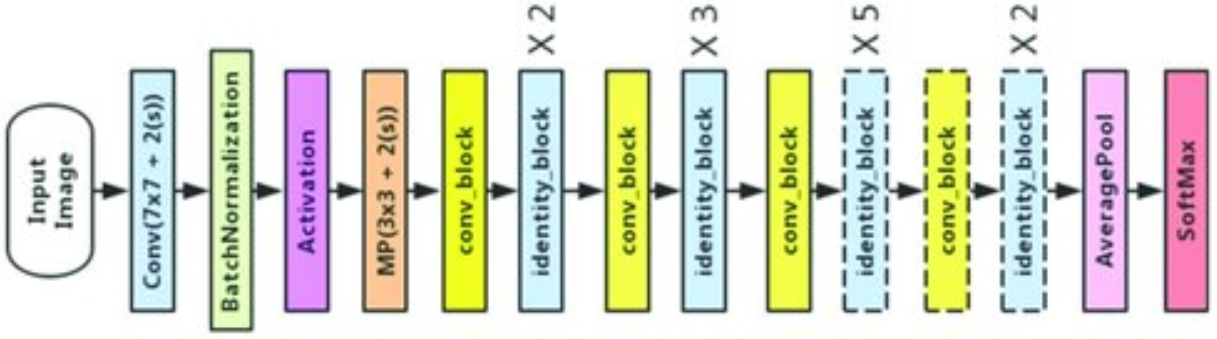}
    \caption{Architecture of ResNet-50 model}
    \label{fig:resnet}
\end{figure}

Fig. 4 shows the architecture of InceptionV3, a deep learning model from the Inception family that is an enhanced version of GoogleNet \cite{b30}. Convolutions and maximum pooling are calculated simultaneously, unlike other architectures like AlexNet \cite{b31} and VGG. To enable deeper networks, it has fewer parameters—under 25 million—than AlexNet, which has 60 million. InceptionV3 has 48 layers and uses 3x3 spatial convolutions.

\begin{figure}[ht]
    \centering
    \includegraphics[width=0.5\textwidth]{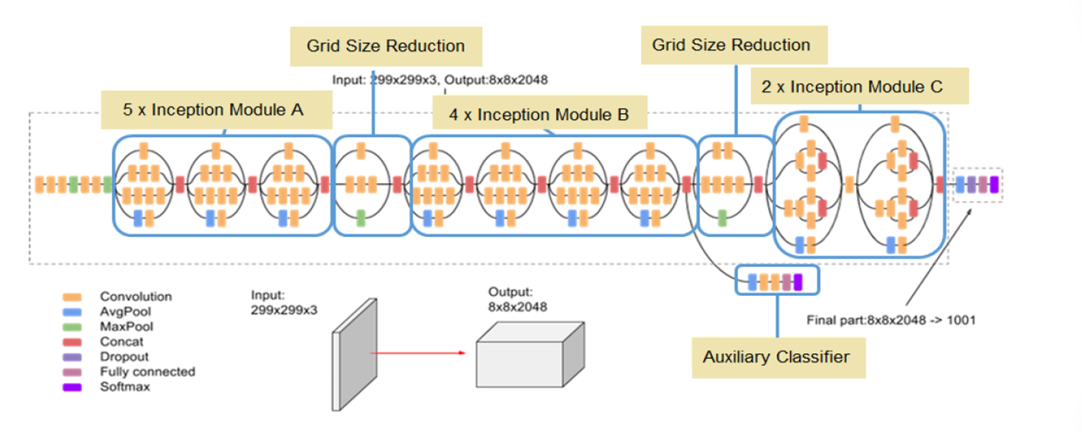}
    \caption{Architecture of InceptionV3 model \cite{b32}}
    \label{fig:inception}
\end{figure}

Xception model, as presented in Fig. 5, is a deep learning architecture that exclusively uses depthwise separable convolution layers. It has 36 convolutional layers and has a comparable parameter count to Inception V3 which lessens the space and time complexity. There is no intermediate ReLU non-linearity, and the order of operations differs from InceptionV3 — 1x1 convolution comes first, followed by spatial convolution.

\begin{figure}[ht]
    \centering
    \includegraphics[width=0.5\textwidth]{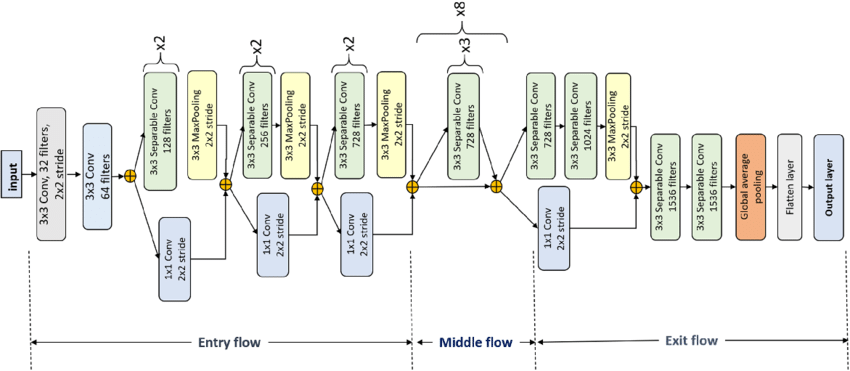}
    \caption{Architecture of Xception model \cite{b33}}
    \label{fig:xception}
\end{figure}

\section{Experimentation and Results}

\subsection{Dataset}\label{AA}
The novel Classroom dataset, captured via a mobile phone camera, aims to identify student behavior in a classroom setting. It includes short 4-5 second video sequences in MP4 format, shot at 30 FPS, and RGB frames with a resolution of 640x480. The focus is on recognizing both usual and unusual student activities, which can be beneficial for enhancing the learning environment. The dataset requires up to 100MB of storage space. Existing video datasets, such as KTH\cite{b34} and Weizman\cite{b35}, primarily consist of simple, usual actions like walking, jogging, and jumping. On the other hand, datasets like UCF101\cite{b36} offer a more complex representation of interactions, focusing on activities like sports or playing music. This novel classroom dataset contains usual and unusual activities observed predominantly in the classroom environment. Fig. 6 displays the activity classes present in the dataset. The activity classes in the dataset are Discussion, Entry/Exit, Hand Raise, Head Down, Talking on mobile, Throwing Objects, and Writing. Real-time recording challenges included dynamic backgrounds, variations in scale and illumination, changing camera views, and capturing multiple people in a single frame, among others. 

\begin{figure*}[ht]
    \centering
    \includegraphics[width=1\textwidth]{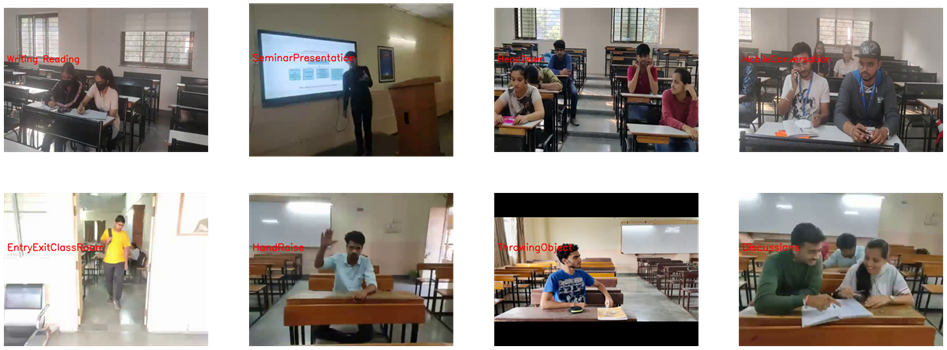}
    \caption{Sample frames from the novel classroom dataset}
    \label{fig:dataset}
\end{figure*}

\begin{figure*}[ht]
    \centering
    \includegraphics[width=0.9\textwidth]{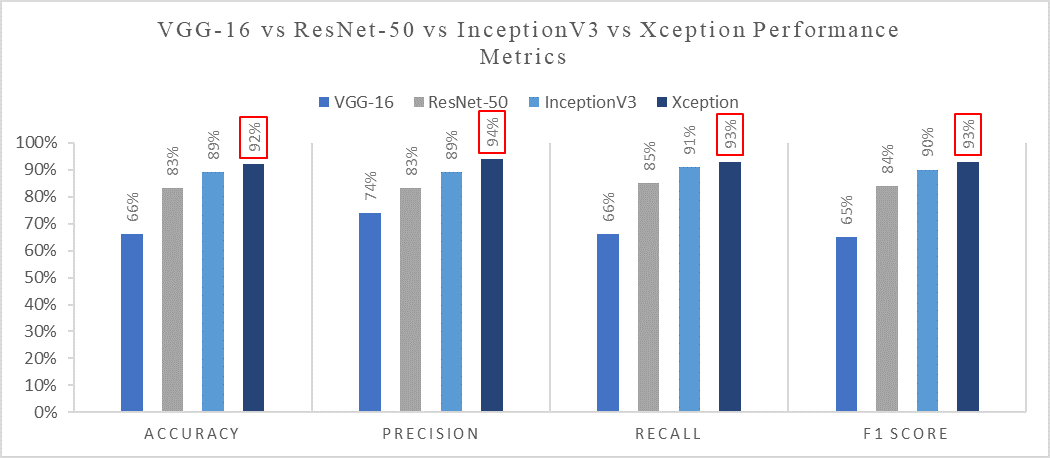}
    \caption{Comparision of the models based on the selected performance metrics}
    \label{fig:comparision}
\end{figure*}

\subsection{Experimental Setup}

This section covers the experimental setup, the process used for training, and the results of the four architectures mentioned previously. The models are tested on a novel classroom dataset recorded and created by the authors, which contains 7 activity classes. Each image has a frame size of 160x160, for a total of 4372 frames. Training, validation, and test data are divided in a ratio of 70:10:20. Adam optimizer having a learning rate of 0.0001 is used along with categorical\_crossentropy loss, and accuracy is employed as an evaluation metric. With a batch size of 8, each model is iterated over 20 epochs.

The base convolution neural network model is generated by disabling the trainable parameters in pre-trained models like VGG16, ResNet50, InceptionV3, and Xception. The newly generated model is succeeded by the Global Average Pooling 2D layer, intended to lower the spatial dimension of data while retaining the essential features. Subsequently, a fully connected layer comprising seven output neurons is connected. The activation function employed here is the softmax function, which is commonly chosen for scenarios involving multi-class classification. Fig. 8 shows the sequential model diagrams and trainable parameters for the proposed model.

\subsection{Performance Measures}
Accuracy, Precision, Recall, and F1 Score are the four performance indicators used in this study, with Accuracy being the most relevant of the four. True positive is indicated by TP, false positive by FP, true negative by TN, and false negative by FN. 

Accuracy is a metric that indicates the model's fraction of exact predictions in relation to the overall number of predictions made. 

\begin{equation}
\text{Accuracy} = \frac{TP + TN}{TP + TN + FP + FN}
\end{equation}

The fraction of true positive predictions to total predicted positive samples is denoted as precision.

\begin{equation}
\text{Precision} = \frac{TP}{TP + FP}
\end{equation}

Recall is defined as the fraction of accurately predicted positive samples among all positive samples.

\begin{equation}
\text{Recall} = \frac{TP}{TP + FN}
\end{equation}

Finally, the F1 score is the combined value of precision and recall, calculated by taking the harmonic mean of precision and recall.

\begin{equation}
\text{F1 Score} = \frac{2*TP}{2*TP + FP + FN}
\end{equation}

\subsection{Results}
As stated previously, 70\% of the dataset is dedicated to training, 20\% to validation, and 10\% is dedicated to testing the models. After employing the models on the testing dataset, a Collation of the four models in terms of output measuring parameters can be observed in Table 1. Xception outperformed the other three models, yielding a testing accuracy of 92\%. Fig. 7 contains a graphical representation of the comparison of the selected models. This study's findings are consistent with the fact that Xception is an improved version of Inception, yielding better results on the novel classroom dataset. The Xception model, due to a depth-wise separable convolutional layer with varied reduced filter dimensionality, improves the accuracy and computation cost.

\begin{figure}[ht]
    \centering
    \includegraphics[width=0.5\textwidth]{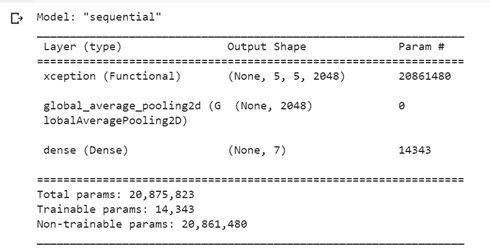}
    \caption{Model Summary}
    \label{fig:model}
\end{figure}

\renewcommand{\arraystretch}{1.5}

\begin{table}[ht]
\centering
\caption{Performance Metrics Comparison}
\label{tab:my-table}
\begin{tabular}{|c|c|c|c|c|}
\hline
\textbf{Model} & \textbf{Accuracy} & \textbf{Precision} & \textbf{Recall} & \textbf{F1 Score} \\ \hline
\textbf{VGG-16} & 66\% &     74\%      &     66\%      &    65\%       \\ \hline
\textbf{ResNet-50} &   83\%       &    83\%         &       85\%      &   84\%          \\ \hline
\textbf{InceptionV3} &   89\%          &      89\%       &    89\%         &   90\%          \\ \hline
\textbf{Xception} &     92\%        &       94\%      &     93\%        &    93\%         \\ \hline
\end{tabular}

\end{table}


The confusion matrix is used to compare predicted and actual class labels to determine the performance of a model, while the Receiver Operating Characteristic (ROC) curve graphically depicts the balance between true positive rate and false positive rate, which aids in the evaluation and comparison of the classification model. The Area Under the Curve (AUC) is the measure of the ability of a classifier to distinguish between classes and is used as a summary of the ROC curve. The classifiers can distinguish well when each class is measured against the others (AUC), but not as well when the prediction probabilities are an output of the softmax function. 

\begin{figure}[ht]
    \centering
    \includegraphics[width=0.5\textwidth]{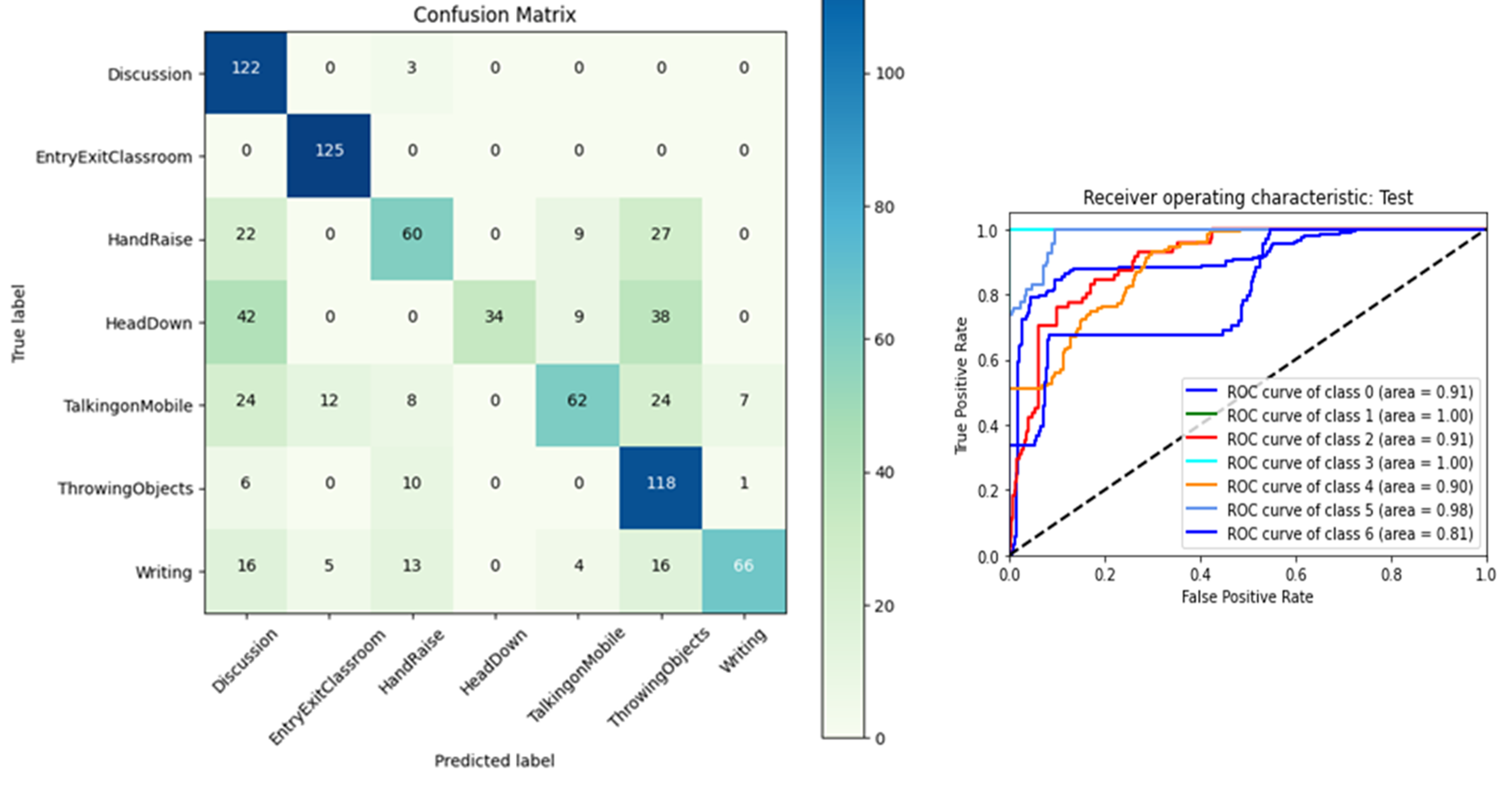}
    \caption{Confusion matrix and ROC Curve for VGG-16 model}
    \label{fig:cmvgg}
\end{figure}

\begin{figure}[ht]
    \centering
    \includegraphics[width=0.5\textwidth]{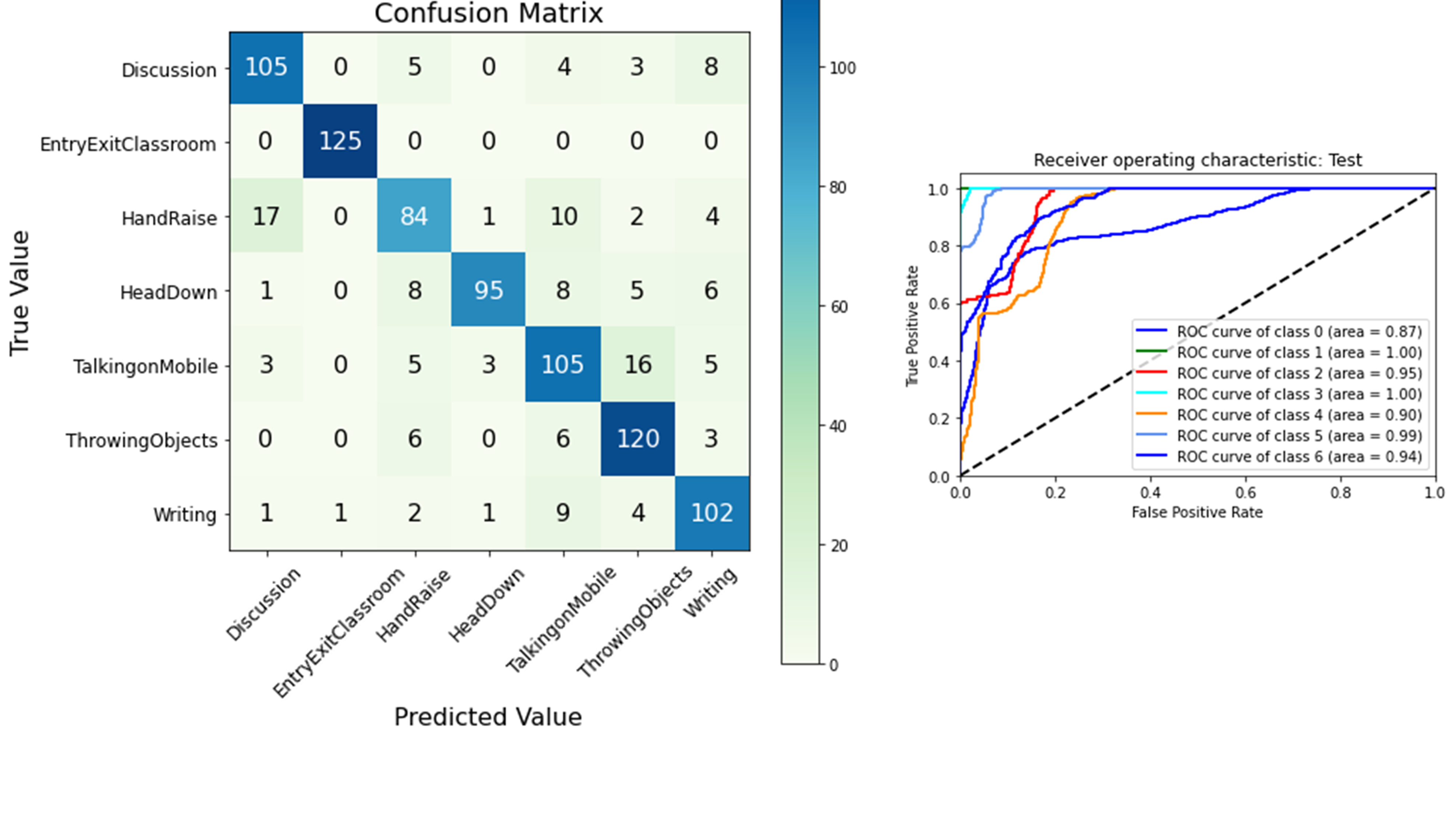}
    \caption{Confusion matrix and ROC Curve for ResNet-50 model}
    \label{fig:cmresnet}
\end{figure}

\begin{figure}[ht]
    \centering    \includegraphics[width=0.5\textwidth]{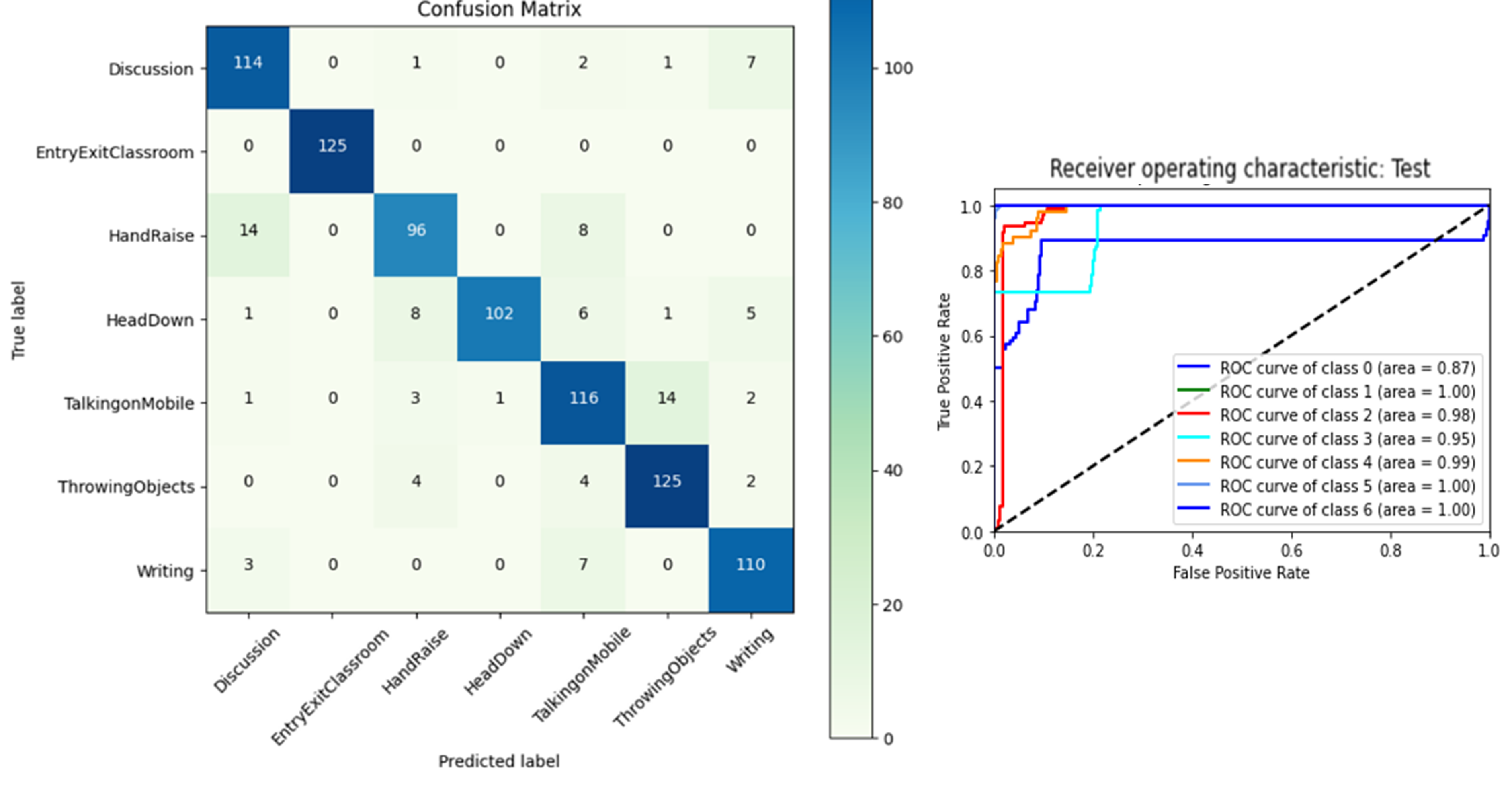}
    \caption{Confusion matrix and ROC Curve for InceptionV3 model}
    \label{fig:cminception}
\end{figure}

\begin{figure}[ht]
    \centering
\includegraphics[width=0.5\textwidth]{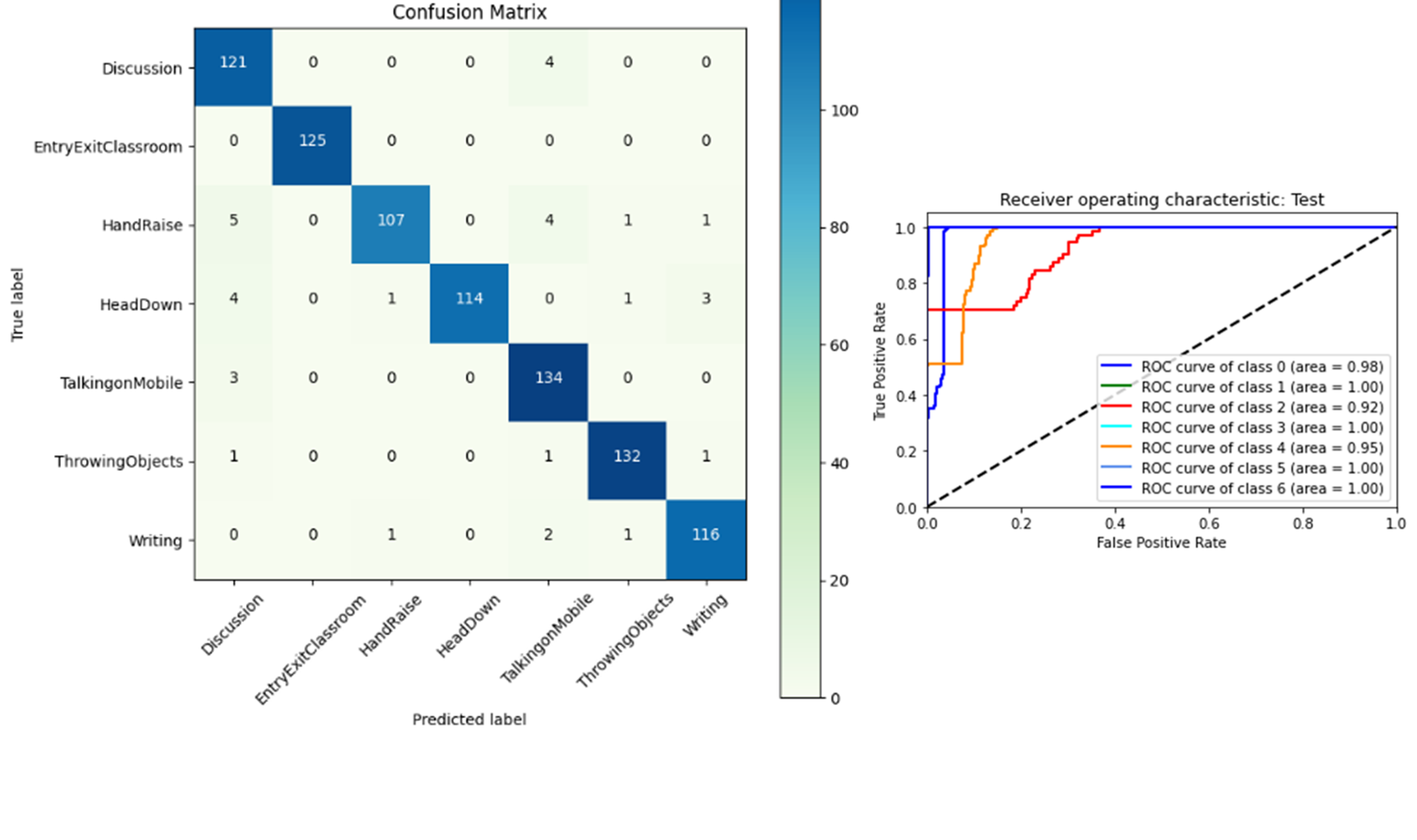}
    \caption{Confusion matrix and ROC Curve for Xception model}
    \label{fig:cmxception}
\end{figure}
\section{Conclusion}
In this paper, we present a Convolutional Neural Network based on transfer learning in solving the problem of student activity recognition using a time series dataset. The research is carried out on a novel classroom dataset comprising over 4,000 images split across 7 activity classes. The developed system is capable of distinguishing seven different student activities, Discussion, Entry/Exit, Hand Raise, Head Down, Talking on mobile, Throwing Objects, and Writing. Various pre-trained convolutional neural networks were studied and four were employed on the novel classroom dataset using the transfer learning approach. The confusion matrices and classification reports show an accuracy of 92\% for the Xception model, followed by 89\% for InceptionV3. The Xception-pertained model supports depthwise separable convolutions, isolates cross-channel and spatial correlations, and captures fine-grained spatial details, which helps in improving recognition accuracy. The ROC curves depict that classifiers are adept at differentiating various classes, but not as well in terms of prediction probabilities, which can be improved by applying superior classifiers. The future focus of this research will be employing the proposed system on a larger and more diverse dataset and the use of cutting-edge models like RNNs and LSTMs for the classification task along with the pre-trained networks.

\end{document}